\documentclass[fleqn]{article}

\usepackage{arxiv}

\usepackage[usenames,dvipsnames]{xcolor}

\usepackage{mypackages}

\bibliography{LLMLink-arXiv-2024}

\usepackage[utf8]{inputenc} 
\usepackage[T1]{fontenc}    
\usepackage{hyperref}       
\usepackage{url}            
\usepackage{booktabs}       
\usepackage{amsfonts}       
\usepackage{nicefrac}       
\usepackage{microtype}      
\usepackage{cleveref}       
\usepackage{lipsum}         
\usepackage{graphicx}
\usepackage{doi}

\usepackage{float}

\title{Predictions from language models for multiple-choice tasks are not robust under variation of scoring methods}

\date{}

\newif\ifuniqueAffiliation
\uniqueAffiliationtrue

\ifuniqueAffiliation 
  \author{ Polina Tsvilodub\thanks{Corresponding author: \texttt{polina.tsvilodub@uni-tuebingen.de}. $\dagger$: equal contributions, listed alphabetically.}$\dagger$, Hening Wang$\dagger$, Sharon Grosch, Michael Franke \\
	Department of Linguistics\\
	University of Tübingen\\
}
\else
\usepackage{authblk}

\author{Michael Franke, Polina Tsvilodub, Fausto Carcassi}
\affil{Department of Linguistics\\University of Tübingen\\
\texttt{[michael.franke|polina.tsvilodub|fausto.carcassi]@uni-tuebingen.de}}
\fi

\renewcommand{\headeright}{}
\renewcommand{\undertitle}{}
\renewcommand{\shorttitle}{Scoring methods for LLM predictions on multiple-choice tasks}

\hypersetup{
pdftitle={Scoring methods for LLM predictions on multiple-choice tasks},
pdfsubject={},
pdfauthor={Polina Tsvilodub, Hening Wang, Sharon Grosch, Michael Franke},
pdfkeywords={large language models, link functions, accuracy, assessment},
}

\newcommand{\tuple}[1]{\langle #1\rangle}

\begin{document}

\maketitle

\begin{abstract}
This paper systematically compares different methods of deriving item-level predictions of language models for multiple-choice tasks.
It compares scoring methods for answer options based on free generation of responses, various probability-based scores, a Likert-scale style rating method, and embedding similarity.
In a case study on pragmatic language interpretation, we find that LLM predictions are not robust under variation of method choice, both within a single LLM and across different LLMs.
As this variability entails pronounced researcher degrees of freedom in reporting results, knowledge of the variability is crucial to secure robustness of results and research integrity.    
\end{abstract}

\section{Introduction}

Recent Large Language Models (LLMs) show impressive performance on various tasks \citep[e.g.,][]{NEURIPS2020_1457c0d6, chowdhery2022palm, openai2023gpt4, touvron2023llama}, leading to discussions about LLMs' status in comparison to human reasoning capabilities or as models of human language \citep[e.g.,][]{bommasani2021opportunities, BinzSchulz2023:Using-cognitive, bubeck2023sparks, Hagendorff2023:Machine-Psychol, katzir2023large, piantadosi2023modern, ShiffrinMitchell2023:Probing-the-psy}. 
These discussions are fuelled by results suggesting excellent model performance, e.g., on large benchmark datasets with multiple-choice answers \citep[e.g.,][]{wang2019superglue, srivastava2022beyond}.
It has been recognized that a more holistic assessment of the performance of LLMs is necessary, which goes beyond mere accuracy and also comprises metrics like bias, efficiency and robustness \citep{LiangBommasani2023:Holistic-Evalua}. 
Similarly, work on calibration of LLM-internal multiple choice answer probabilities with respect to accuracy of factual knowledge highlights the importance of careful assessment of robustness of LLMs \citep[e.g.,][]{kadavath2022language, 10.1162/tacl_a_00494}.
However, comparatively little work directly compares different methods for deriving LLM predictions for benchmark testing or in relation to human behavior \citep[but see, e.g.,][]{PoliakNaradowsky2018:Hypothesis-Only,holtzman-etal-2021-surface,hu2023prompt,tedeschi-etal-2023-whats}. 

This paper contributes to previous work on robustness, by exploring possible variance at the level of an LLM's prediction for a \textit{single item} of a multiple choice task.
The focus is on a pragmatic language-understanding task \citep{hu-etal-2023-fine}, so as to confine the investigation to a single domain in which language models should, arguably, show high performance.
Indeed, experimental work on human language understanding shows that different \textit{task types}, i.e., ways of asking a question, can lead to qualitatively different patterns in human responses \citep[e.g.,][]{GeurtsPouscoulous2009:Embedded-Implic,ChemlaSpector2010:Experimental-Ev}, thereby stressing the importance of investigating different \textit{linking functions}, which map model-internal information to aspects of human data \citep{DegenGoodman2014:Lost-your-Marbl,Franke2016:Task-types-link,JasbiWaldon2019:Linking-Hypothe}. 
Taking inspiration from this literature, we formulate a range of methods for retrieving LLM answers in multiple-choice tasks, both for accuracy scoring and for assessing human-likeness of LLMs' predictions. 
In particular, we build on three methods commonly used in psycholinguistic experiments: free production of answers, forced choice answer selection, and Likert-scale rating of answers.
Additionally, we consider link-functions based on embedding similarity.\footnote{All materials can be found under \url{https://tinyurl.com/4czu7mbu}.}
 

\section{Related work}

Recent like-minded work has analysed the human-likeness of LLMs' linguistic performance, e.g., testing language models on different syntactic phenomena \citep{wilcox2018rnn, wilcox2020onlineprocessing, futrell-etal-2019-neural, arehalli-etal-2022-syntactic}. Human-likeness of LLM performance has also been assessed on other domains like semantic judgements \citep[e.g.,][]{levy2017semantic, kauf2022event}. 
Other work has focused on comparing variability of texts generated by LLMs to human production variability \citep[e.g.,][]{meister-cotterell-2021-language, giulianelli2023comes}.  
Most relevantly to the present study, recent work has looked at the human-likeness of LLM performance on subtle pragmatic phenomena like irony or compliance with Gricean maxims \citep{hu-etal-2023-fine, tsvilodub2023overinformative,San-PietroFrau2023:The-pragmatic-p}. However, these comparisons typically assess human-likeness based on one particular LLM answer scoring method (i.e., one linking function).

Akin to variability in human experimental work, different methods of retrieving LLM predictions are common. 
There are abundant approaches to generating answers from LLMs by many sophisticated prompting strategies \citep[e.g.,][among many others]{nye2021show, wei2022chain, yao2023tree}. 
Next to directly sampling answers from the LLMs, surprisals or log probabilities of tokens are widely used \citep[e.g.,][]{wilcox2020onlineprocessing, futrell-etal-2019-neural, kauf2022event, hu-etal-2023-fine}, which are often aggregated in different ways \citep{holtzman-etal-2021-surface}.
Further, cosine similarity is commonly applied in NLP; it has also been used to relate distributional word representations to human performance in lexical priming studies (\citet{doi:10.1080/17470218.2015.1038280}, but see \citet{zhou-etal-2022-problems} for a critical discussion). 
 
However, these methods used in different studies (oriented towards comparison to human performance or NLP-typical computation of aggregated performance measures) have not often been compared on the same task, or for different models. 
Therefore, the present study systematically compares different methods used in previous work in a controlled setting, highlighting parallels to psycholinguistic experiments and comparison of LLM performance to human behavior.

\section{Experimental Set-up}
\label{sec:append-how-prod}

\subsection{Materials, Notation \& Models}
\label{section:materials-procedure}

We use the material and human data from a forced-choice experiment on pragmatic language interpretation provided by \citet{hu-etal-2023-fine}. 
Seven experimental conditions were tested, targeting different phenomena of interest: coherence, humor, deceit, irony, indirect speech, Gricean maxims and metaphors.
There are 20 to 40 items per condition. 
Each item consists of a context and a trigger sentence. 
Human participants indicated their interpretation of the trigger by selecting a multiple-choice answer option.
Each vignette had two, four or five options, one of which is the \textit{target option} representing a pragmatic, non-literal interpretation.
\begin{figure*}[t!]
\centering
    \includegraphics[width=0.45\linewidth]{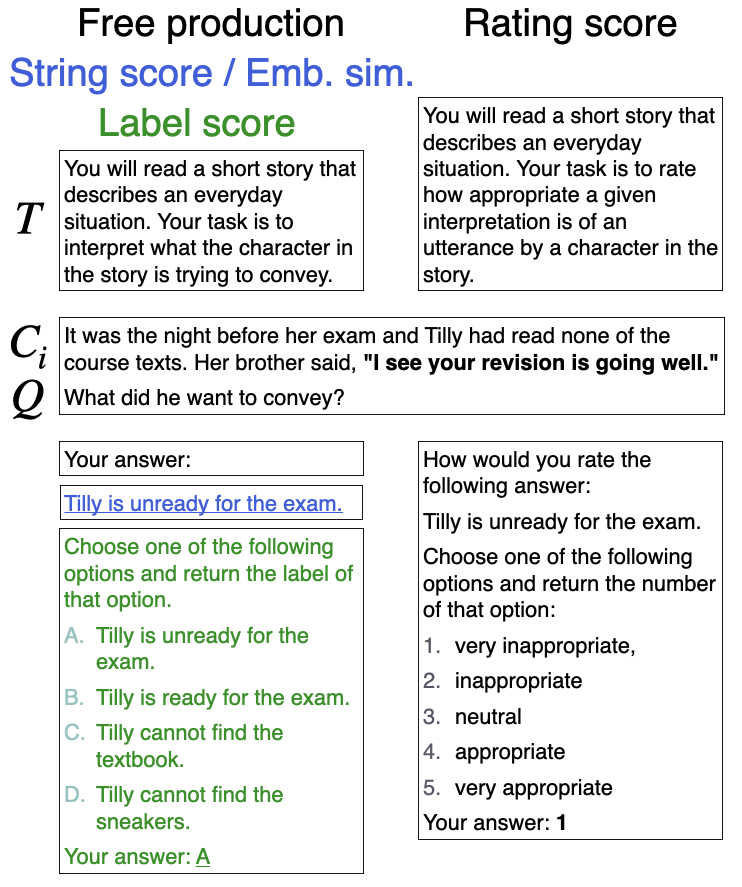}
  \hfill
    \raisebox{.1\height}{\includegraphics[width=0.45\linewidth]{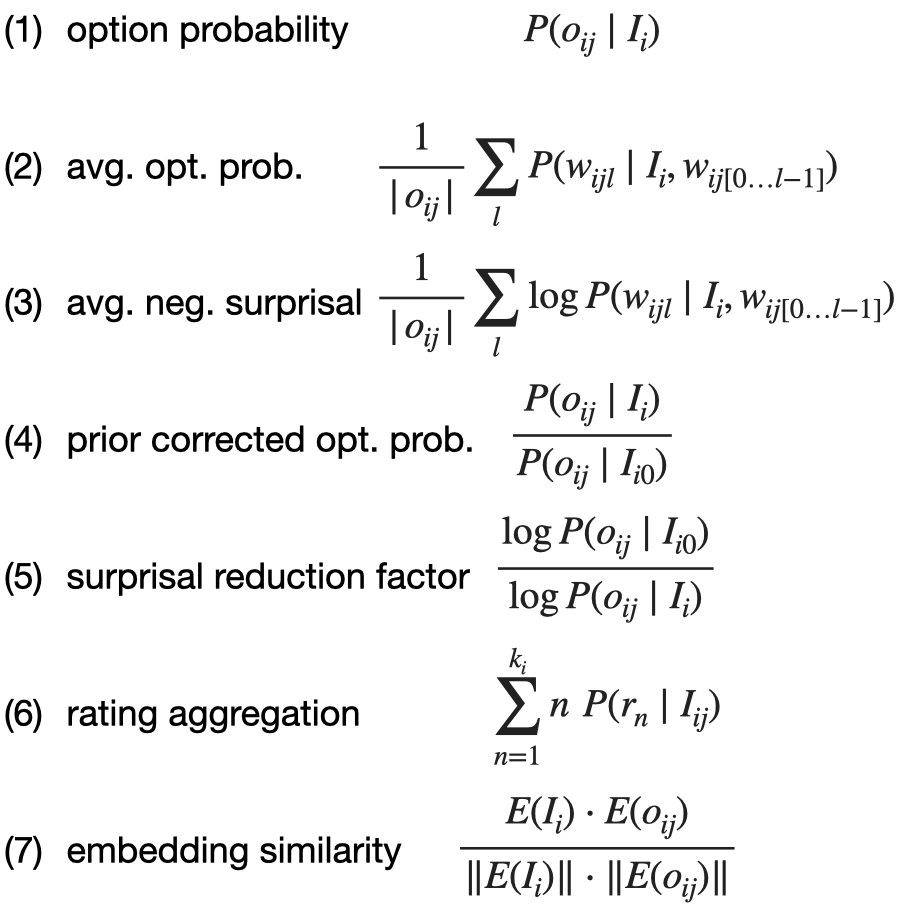}}
\caption{\textbf{Left:} example of stimulus material (irony condition) for different methods.  Sequences for which scores are retrieved are underlined. The trigger sentence is in boldface. $T$, $C_i$ and $Q$ are included in all methods' input prompts $I_i$. Colors indicate the text which is additionally used for the respective method. 
\textbf{Right:} set of all relevant scores. \label{fig:scores-overview}}
\end{figure*}


Our experimental material was derived from that of \citet{hu-etal-2023-fine} as follows.
All items of the same condition shared the task description $T$ and question $Q$ (see Figure~\ref{fig:scores-overview}, left).
Instructions and prompts differed slightly between phenomena (e.g., the coherence condition asked to decide whether a pair of sentences was coherent; the maxims condition asked what the speaker of the sentence was trying to convey).
Each experimental item $i$ consisted of a context description $C_{i}$, including the trigger sentence, and $k_i$ answer options $\tuple{o_{ij}}_{{1 \le j \le k_{i}}}$, with $o_{ij} = \tuple{w_{ijl}}_{1 \le l \le |o_{ij}|}$ being a sequence of words.

We used four LLMs for text generating and probability assignement to text: GPT-3.5-turbo-instruct and text-davinci-002 models (175B parameters) \citep{NEURIPS2020_1457c0d6, ouyang2022training}, LLaMA-2 (7B parameters) \citep{touvron2023llama}, and FLAN-T5-XL (3B parameters) \citep{chung2022scaling}. 
To retrieve token probabilities form FLAN-T5-XL, which is an encoder-decoder model partially pretrained with a denoising objective, we used the pseudo-likelihood computation \citep{salazar-etal-2020-masked}. That is, we created inputs consisting of the $T$, $Q$, $C_i$ and $\tuple{w_{ij1}, w_{ij2}, \dots , w_{ijk-1}, w_{ijk+1}, \dots, w_{ijl}}$ where the answer option word $w_{ijk}$ was masked with the special token \texttt{$\langle$extra\_id\_0$\rangle$}. We created outputs of the form \texttt{$\langle$extra\_id\_0$\rangle$} $w_{ijk}$ \texttt{$\langle$extra\_id\_1$\rangle$} and summed the retrieved log probabilities of the tokens corresponding to $w_{ijk}$ (with only one mask for sentence-initial and final tokens).\footnote{At each scoring step we masked tokens corresponding to an entire word. This is closely related to work by \citet{kauf-ivanova-2023-better}. Exploratory work masking single (subword) tokens revealed worse performance under string scoring and, therefore, is not reported.}
We also used text-embedding-ada-002 for the embedding similarity method \citep{Greene_Sanders_Weng_Neelakantan_2022}.

\subsection{Methods of Answer Selection}
\label{sec:methods-scores}

We compare five different methods of determining an LLM's answer choice: 
(i) free generation, 
(ii) string scoring, 
(iii) label scoring, 
(iv) rating aggregation, and 
(v) embedding similarity.
All methods, except for free generation, select the answer with the highest numerical score.
For some methods (string and label scoring), several candidate scores are compared.
Figure~\ref{fig:scores-overview} (right) lists all scores used.

\subsubsection{Free Generation}

The free-generation method prompts an LLM to generate continuations, which were then manually classified as correct (corresponding to the target option) or not by the authors.
The input prompt $I_{i}$ for item $i$ can be seen in Figure~\ref{fig:scores-overview} (left).
Generations of at most 50 tokens were obtained by sampling temperatures $\tau=0.1$ and $\tau=0.9$, with five random seeds, respectively. There were no qualitative differences between the results obtained with different parameters parameters, so only results generated with $\tau=0.1$ are reported.

\subsubsection{String Scoring}
\label{section:string-scoring}

The string-scoring method assigns a numerical score to each answer option $o_{ij}$ of item $i$.
We compute the \textit{option probability}, i.e., the conditional probability of $o_{ij}$ given $I_i$ (Fig.~\ref{fig:scores-overview}, right, (1)).
Following common practice \citep[e.g.,][]{bommasani2021opportunities}, we also consider length-corrected measures, \textit{average option probability} (2) and \textit{average negative surprisal} (3). 
Finally, to compensate for different baseline probabilities of different options, we compute a \textit{prior corrected option probability} (4), where $I_{i0}$ is like $I_i$ but without $C_i$ and $Q$ \citep[cf.,][]{PoliakNaradowsky2018:Hypothesis-Only, holtzman-etal-2021-surface}.
For symmetry, we also compute the \textit{surprisal reduction factor} (5), a prior correction for surprisals.

\subsubsection{Label Scoring}

While the string-scoring method presented each option $o_{ij}$ individually for scoring, the label-scoring method presents all options with a uniquely identifying label in the input prompt, and computes scores from the probability assigned to the labels (see Figure~\ref{fig:scores-overview}, left).
Let $\tuple{l_{ij}}_{{1 \le j \le k_{i}}}$ be  a set of ordered answer labels (e.g., A, B, ... or 1, 2, ...) for the $k_i$ options of item $i$. 
We consider the \textit{label probability} $P(l_{ij} \mid I_i)$ as the analogon of score (1).
Since the analogues of length-corrected scores (2) and (3) for labels are equivalent to (1), we only consider scores of type (1), (4) and (5) for labels.

\subsubsection{Rating Aggregation}
The rating-aggregation method computes a score via a procedure akin to Likert-scale rating tasks in human experiments.
For each answer option $o_{ij}$, the LLM is asked to select a rating from a five-point scale with labels $r_1, \dots, r_5$, which are internally treated as a number, $1, \dots, 5$.
The \textit{rating aggregation} score assigned to option $o_{ij}$ is the probability-weighted average of the rating (see Figure~\ref{fig:scores-overview}, right, (6)).
The experiments used rating scales based on randomly chosen concepts: ``likely'', ``appropriate'', ``plausible'', ``possible''.
Reported results are based on the best-performing scale for each model.

\subsubsection{Embedding Similarity}

Finally, motivated by the common use of information extracted from embeddings, the embedding-similarity method compares options based on an \textit{embedding similarity score} (7) between input prompt embeddings $E(I_{i})$ and embeddings $E(o_{ij})$ of each answer option $o_{ij}$. 
The input prompt $I_{i}$ for item $i$ is constant and identical to the string score inputs (see Figure~\ref{fig:scores-overview}, left). 
For LLaMA-2, we use the last hidden representation of the last sequence token as input and answer option representations, respectively. 
For FLAN-T5, we use the last hidden representation of the last token of the encoder as the input embedding, and the last hidden representation of the decoder conditioned on the input for the answer option embedding \citep[cf.][]{ni-etal-2022-sentence}. 
For the OpenAI models, we use sequence embeddings by \citet{Greene_Sanders_Weng_Neelakantan_2022}. 

\begin{figure*}[t]
    \centering
    \includegraphics[width=\textwidth]{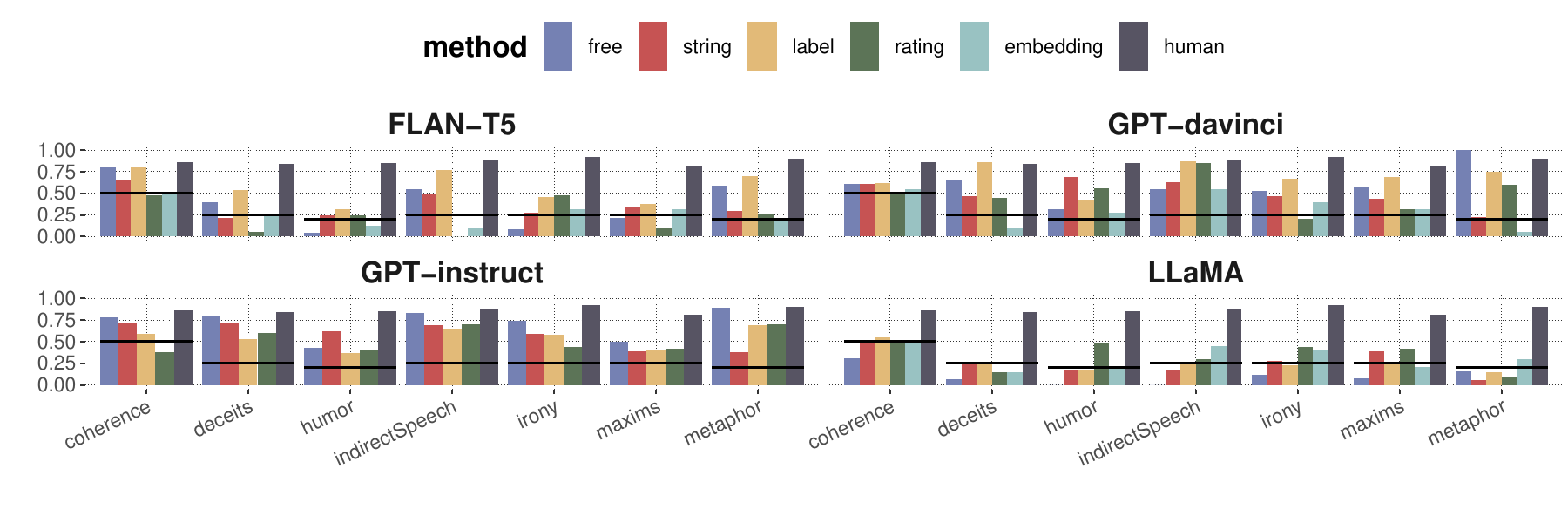}
    \caption{Task accuracy for different methods and models, separately for different conditions.
    Bars show accuracy scores, which are averages over different versions of each method (e.g., random seeds or different scores).
    Horizontal bar shows accuracy expected from random guessing in each condition.
    }
    \label{fig:results-byPhenomenon}
\end{figure*}

\section{Results}
\begin{figure*}[t]
    \centering
    \includegraphics[width=\textwidth]{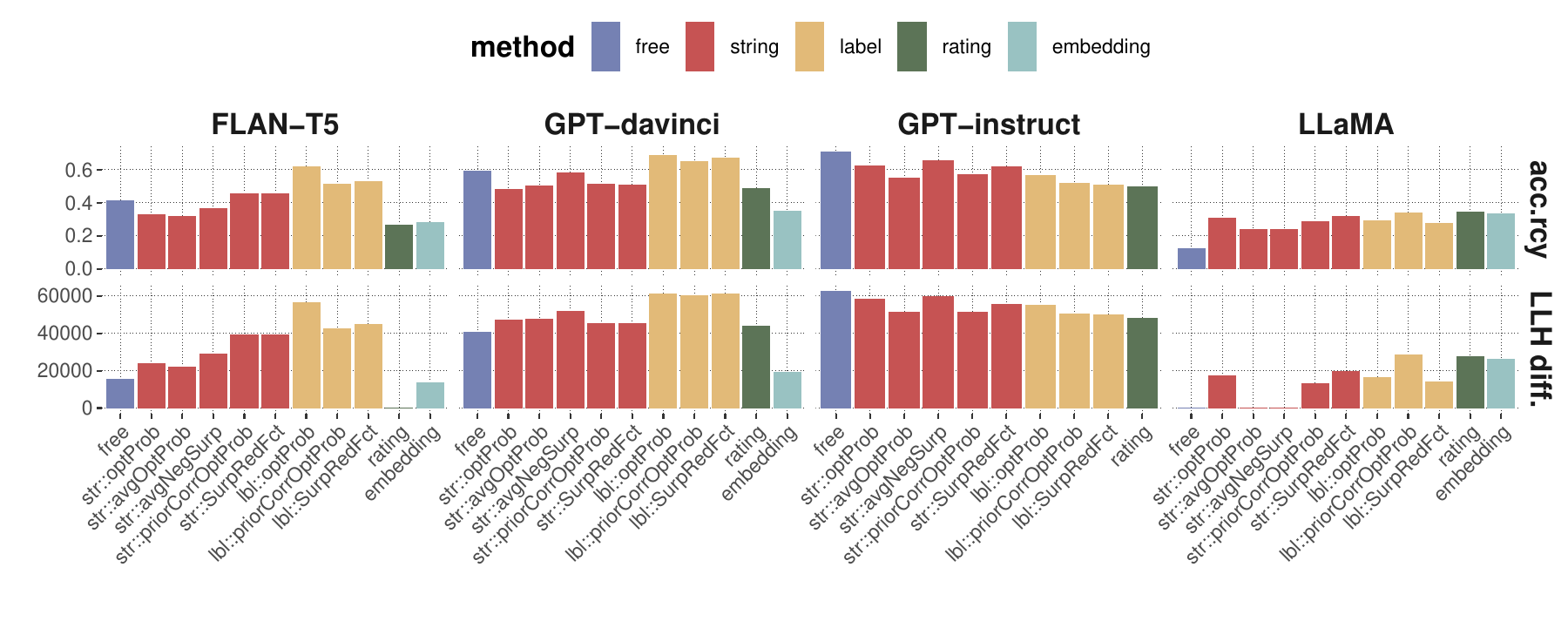}
    \caption{
    Results of methods comparison, based on accuracy (proportion of target choices) and goodness-of-fit to the human data. Different scores are displayed on the x-axis.
    For ease of visual comparison, goodness-of-fit is shown as difference in log-likelihood of the human data under the LLM predictor, compared against one of the worst performing models (higher is better).}
    \label{fig:results-combined}
\end{figure*}


Figure~\ref{fig:results-byPhenomenon} shows accuracy (proportion of target choices) for different models, methods and conditions.
Accuracy is averaged over different scoring functions, except where indicated otherwise in the description of the methods in Section~\ref{sec:methods-scores}.
Though human accuracy (dark gray bars) is consistently high, the performance of models and methods varies considerably, even qualitatively, across different conditions: for example, for GTP-davinci string-scoring, label-scoring and generation-based methods take turns outperforming each other across conditions. 

Figure~\ref{fig:results-combined} shows, for each model, method, and, if applicable, score, the accuracy, as well as a measure of goodness-of-fit to human data (i.e., log-likelihood of human data), averaged over conditions.
The log-likelihood of human data is the sum of log-probabilities, assuming the obvious binomial likelihood function, under an LLM-based point-valued, probabilistic prediction $p$ for each phenomenon, model and score in the data set. 
The target rate $p$ is the proportion of target choices for any given triple of phenomenon, model and metric.

We observe that accuracy and likelihood of data are fairly correlated. 
This is not logically necessarily, but explainable in the present case, as there is no model-method pair which outperforms human accuracy in all conditions (see Figure~\ref{fig:results-byPhenomenon}).
This correlation permits occasional glossing over differences between accuracy and likelihood.

There are pronounced performance differences between different models.
LLaMA-7B performed rather poorly, especially for free-generation and string-scoring methods.
Models from the GPT-family performed best.
GPT-instruct with free-generation achieved the highest accuracy and likelihood across all models, methods and scores.

There is no strong evidence for a single method delivering supreme results for all models.
Label scoring worked best for all models except for GPT-instruct, where free generation achieves slightly higher accuracy and likelihood and where even string-scoring outperforms label-scoring.
Judging from visual inspection, the choice of method (and score) seems to matter more for models which overall perform worse than for the best performing models.
This is unlikely a mere ceiling effect for the well-performing models, since accuracy is far from ceiling even for the best performing model.

Based on the obtained results, we can recommend \emph{not} using rating approaches, because they are more expensive to compute with no obvious payback in performance or goodness-of-fit.
The embedding-based method also performed generally worse.
Free generation gave excellent results for GPT-instruct, which may be due to the fact that it is highly optimized to give coherent generation, thus leading to fewer off-topic responses.
The improvement under free generation over scoring for GPT-instruct, which was fine-tuned with RLHF, could also be because log-probability calibration might deteriorate upon fine-tuning \citep[cf.][]{openai2023gpt4}.

\section{Discussion and Conclusion}
We presented a case study comparing different methods of assessing performance of LLMs, both in terms of accuracy (matching a gold-standard) and goodness-of-fit to human data. 
Methods we considered were standard from the previous literature and/or inspired by different task types for assessing human decisions in multiple-choice contexts.
Best and most stable results were achieved with label scoring, in line with previous results \citep[cf.][]{hu2023prompt}. 
We found that different methods of assessing model performance give different results for different models and conditions. 
This calls for attention to variability in performance assessment.

Awareness of variability in performance assessment is important for good practices in reporting and building on results.
Variability of this kind provides researcher degrees of freedom, which can lead to accumulation of false or overstated results and biased reporting, even in the absence of misleading intentions \citep{Ioannidis2005:Why-Most-Publis,Chambers2017:The-Seven-Deadl}.
To reduce researcher degrees of freedom and to build more robust and reproducible research practices \citep{WielingRawee2018:Reproducibility}, preregistered analyses are an option \citep{MunafoNosek2017:A-manifesto-for}.

The present investigation has a narrow empirical scope, by focusing on a single data set from a single domain.
Strictly speaking, this is not detrimental to the demonstration of the general point that results suffer from non-robustness under available researcher degrees of freedom in determining an assessment method.
Nevertheless, future like-minded comparisons of methods should extend to other multiple-choice datasets and tasks where human performance results are available, to tokenizer architectures other than BPE, and other LLM configurations like different decoding schemes for free generation. 
Additionally, future work should go beyond choice of a single target option and address full distributional predictions for the whole set of options. 
Assessment of distributional predictions matters to understand the tail-behavior of generation, e.g., when sampling with non-zero temperature. 
Finally, to better understand the sources of variability across methods, effects of factors like LLM size, architecture, and training data could be investigated. 

\section{Limitations}
The reproducibility of results under closed-source models like GPT-3.5-turbo-instruct and text-davinci-002 is limited because the respective API endpoints were disabled by OpenAI after this work was completed. Furthermore, these results are limited to a small set of manually created experimental items from prior work \citep{hu-etal-2023-fine}. The items are only in English, and test pragmatic phenomena in a way which is specific to Western social norms and culture \citep{rubio2020incrementality}. Future work should extend such evaluations to other languages. Finally, many of the tested pragmatic phenomena and items might be confounded with effects of politeness, so future work could extend to testing LLM performance on politeness datasets directly \citep[e.g.,][]{yoon2020polite, carcassi2023truth}.

\section{Acknowledgments}
We gratefully acknowledge support by the state of Baden-W\"urttemberg, Germany, through the computing resources provided by bwHPC and the German Research Foundation (DFG) through grant INST 35/1597-1 FUGG. Michael Franke is a member of the Machine Learning Cluster of Excellence, EXC number 2064/1 – Project number 39072764.

\printbibliography[heading=bibintoc]

\end{document}